\documentclass[letterpaper]{article} 
\usepackage{aaai2026}  
\usepackage{times}  
\usepackage{helvet}  
\usepackage{courier}  
\usepackage[hyphens]{url}  
\usepackage{graphicx} 
\usepackage{natbib}  
\usepackage{caption} 
\usepackage{soul}
\usepackage{url}
\usepackage[utf8]{inputenc}
\usepackage{caption}
\usepackage{booktabs}
\usepackage{algorithm}
\usepackage{algorithmic}
\usepackage[switch]{lineno}
\usepackage{pifont}

\usepackage{amsmath,amssymb,amsfonts}
\usepackage{textcomp}
\usepackage{multirow}
\usepackage{subfigure}
\usepackage{array}
\usepackage{colortbl}
\usepackage{xcolor}
\usepackage{mathrsfs}
\usepackage{algorithm}
\usepackage{algorithmic}

\usepackage{newfloat}
\usepackage{listings}
\DeclareCaptionStyle{ruled}{labelfont=normalfont,labelsep=colon,strut=off} 
\floatstyle{ruled}
\newfloat{listing}{tb}{lst}{}
\floatname{listing}{Listing}
\title{Perspective from a Broader Context: Can Room Style Knowledge Help Visual Floorplan Localization?}
\author{
    Bolei Chen,
    Shengsheng Yan,
    Yongzheng Cui,
    Jiaxu Kang,
    Ping Zhong\thanks{Corresponding author.},
    Jianxin Wang$^{*}$
}
\affiliations{
    \textsuperscript{\rm }School of Computer Science and Engineering, Central South University\\

    \{boleichen, shengshengyan, yongzhengcui, jxkang, ping.zhong\}@csu.edu.cn, jxwang@mail.csu.edu.cn
}

\usepackage{bibentry}

\begin{document}

\maketitle

\begin{abstract}
Since a building's floorplan remains consistent over time and is inherently robust to changes in visual appearance, visual \textbf{F}loorplan \textbf{Loc}alization (FLoc) has received increasing attention from researchers. However, as a compact and minimalist representation of the building's layout, floorplans contain many repetitive structures (e.g., hallways and corners), thus easily result in ambiguous localization. Existing methods either pin their hopes on matching 2D structural cues in floorplans or rely on 3D geometry-constrained visual pre-trainings, ignoring the richer contextual information provided by visual images. In this paper, we suggest using broader visual scene context to empower FLoc algorithms with scene layout priors to eliminate localization uncertainty. In particular, we propose an unsupervised learning technique with clustering constraints to pre-train a room discriminator on self-collected unlabeled room images. Such a discriminator can empirically extract the hidden room type of the observed image and distinguish it from other room types. By injecting the scene context information summarized by the discriminator into an FLoc algorithm, the room style knowledge is effectively exploited to guide definite visual FLoc. We conducted sufficient comparative studies on two standard visual Floc benchmarks. Our experiments show that our approach outperforms state-of-the-art methods and achieves significant improvements in robustness and accuracy.

\end{abstract}


\section{Introduction}

Camera localization is a long-standing problem in computer vision, widely used in 3D reconstruction \cite{liu2017efficient}, AR/VR applications of mobile devices, and robotic navigation \cite{li2024flona}. Due to the complex room layouts and absence of satellite location signals, visual localization in indoor scenes is particularly challenging. Classical visual localization algorithms rely heavily on pre-collected databases \citep{balntas2018relocnet, 2017NetVLAD} or complex 3D scene reconstructions \citep{liu2017efficient, sarlin2019coarse, sattler2016efficient}, which are expensive to build, store and maintain. Since floorplans are lightweight, easily accessible, consistent over time, and inherently robust to changes in visual appearance, some recent work has explored the problem of localizing camera observations by matching depth-based structural cues in the provided 2D floorplan. Such floorplans, which can be found in places such as school buildings, train stations, and apartments, encode rich and sufficient information to aid visual localization in unvisited scenes. 

\begin{figure}[!t]
 \centering
 \includegraphics[width=1.0\linewidth]{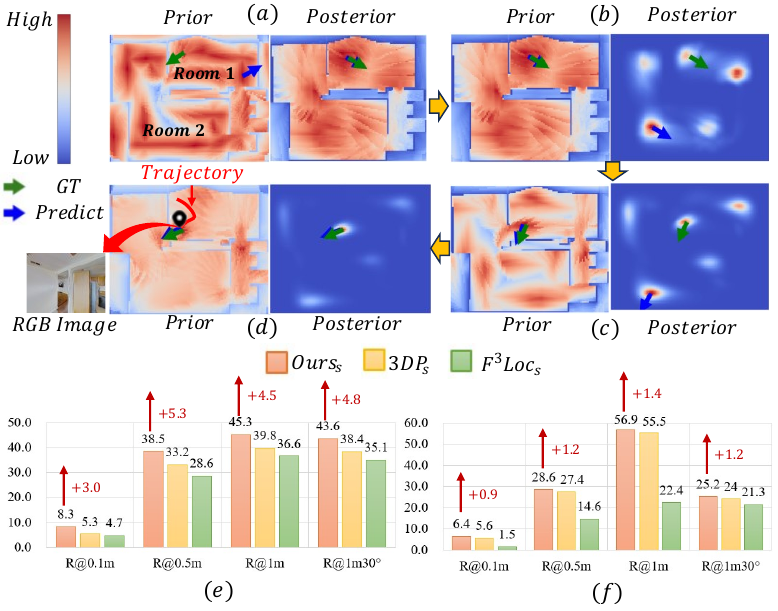}
 \caption{(a)-(d) show the ambiguous localizations in F$^3$Loc method \cite{chen2024f3loc}. $Prior$ denotes the probabilistic map of a single-frame visual Floc based on the current visual image. $Posterior$ denotes the probabilistic map of a long-sequence visual FLoc using Bayesian filtering. Red color indicates high likelihood. (e) and (f) show our method significantly improves the Floc accuracy relative to the SoTA methods (3DP \cite{chen20253dp} and F$^3$Loc) on the challenging Gibson and Structured3D datasets, respectively.}
 \label{fig1}
 \vspace{-0.3cm}
\end{figure}

Existing methods \cite{karkus2018particle, chen2024f3loc} typically match visual features with 2D geometric cues in floorplans to achieve visual \textbf{F}loorplan \textbf{Loc}alization (FLoc). However, as a compact representation of the building’s layout, floorplans contain many repetitive structures, thus easily result in ambiguous localization. As shown by the prior probabilistic map in Figure \ref{fig1} (a), Room 1 has repetitive corner structures, leading to incorrect localization in a single room. In addition, the layouts of Room 1 and Room 2 are very similar, which can easily mislead the visual FLoc method to localize itself in the wrong room, as shown by the posterior probabilistic maps in Figure \ref{fig1} (b) and (c). To address these issues, existing methods make efforts in 2D semantic/3D geometric priors exploitation \citep{min2022laser, chen20253dp, grader2025supercharging} and Bayesian filtering-based sequential localization \citep{karkus2018particle, chen2024f3loc}. 

On the one hand, the \textbf{S}tate-\textbf{o}f-\textbf{T}he-\textbf{A}rt (SoTA) method \cite{chen20253dp} injects 3D geometric priors into a visual FLoc algorithm through unsupervised learning, significantly improving FLoc accuracy. However, 3D geometric cues can not effectively resolve localization ambiguities caused by misleading scene layouts. Some other methods \cite{min2022laser, grader2025supercharging} utilize additional room category annotations or semantic labels in the floorplan (e.g., windows and doors) to assist visual FLoc. However, these semantics are not always available in floorplans or require costly manual annotations as supervised signals. On the other hand, since image sequences \citep{sarlin2022lamar, sarlin2023orienternet} can somewhat eliminate localization uncertainty, other methods \cite{karkus2018particle, chen2024f3loc} use Bayesian filtering to optimize the posterior distribution of the current pose. However, we experimentally find that only when there is a significant change in visual appearance (e.g., when about to exit a room) can localization ambiguity be effectively alleviated, as shown by the probabilistic maps in Figure \ref{fig1} (d). That is, image sequences can only mitigate FLoc ambiguity to a limited extent.

In this work, we propose a principled method to mitigate localization uncertainty caused by repetitive or similar structures by utilizing broader scene contextual information. This means that we have to figure out what we can rely on to infer indoor room relationships when only RGB images are available. We observe that different indoor rooms, such as bedrooms, bathrooms, and kitchens, typically have their own specific styles, such as decorative styles and furniture. These variations are mainly due to the different functions and requirements of each room. Therefore, visual FLoc algorithms can potentially identify room styles from current visual signals and distinguish them from other room styles. 

Technically, we propose an clustering-constrained unsupervised learning technique to train a room discriminator on a self-collected unlabeled RGB image dataset. The RGB images are collected using an automated pipeline based on publicly available indoor scene datasets \cite{xia2018gibson} and corresponding robotic navigation datasets \cite{mezghan2022memory}, including images from different angles within the same room and images across rooms. The well-trained room discriminator can empirically extract hidden room types from observed images and explicitly distinguish them from other room types. Our visual FLoc method is built upon F$^3$Loc \cite{chen2024f3loc} and consists of a front-end observation model and a back-end histogram filter. By injecting the scene layout priors summarized by the discriminator into the observation model, the room style knowledge is effectively exploited to guide definite visual FLoc. We conducted sufficient comparative studies on two standard visual Floc benchmarks to evaluate our method. Our method achieves SoTA visual FLoc performance and significantly outperforms the strong baselines, as shown in Figure \ref{fig1} (e) and (f). Overall, our main contributions are as follows:

(1) We discuss the feasibility of learning indoor room relationships from RGB images and propose a principled solution to alleviate the localization ambiguity of visual FLoc.

(2) To model room style knowledge, a clustering-constrained unsupervised learning technique is proposed to train a room discriminator on an automatically collected unlabeled RGB image dataset.

(3) By injecting room style knowledge summarized by the discriminator into a visual FLoc algorithm, our method achieves SoTA FLoc performance.

\section{Related Work}

\subsection{Floorplan Localization}

FLoc tasks are often associated with LiDAR-based \textbf{M}onte \textbf{C}arlo \textbf{L}ocalization (MCL) \citep{dellaert1999monte, chu2015you, mendez2018sedar, winterhalter2015accurate}, which is a classical framework for 2D localization on purely geometric maps. However, the usage of LiDAR hinders the application of such localization algorithms on common mobile devices. To alleviate this limitation, some work \citep{boniardi2019robot, chu2015you, howard2022lalaloc++, howard2021lalaloc, min2022laser} investigate visual FLoc based on monocular and panoramic images. Some of these methods leverage 2D scene priors \citep{boniardi2019robot} and visual features \citep{min2022laser} by matching them with scene layouts to achieve visual FLoc. Several other methods \citep{howard2022lalaloc++, howard2021lalaloc} localize by comparing the panoramic image features rendered at specific locations with the query image features. However, these methods either assume known camera and room heights or require panoramic images, which limits the generalization of the localization algorithms. 

Recently, researchers have been working on generic monocular vision FLoc techniques \citep{karkus2018particle, chen2024f3loc, chen20253dp} that employ Bayesian filters \citep{jonschkowski2016end, bishop2001introduction} to solve the long-sequence FLoc problem. Despite promising progress, these methods suffer from localization uncertainty caused by repetitive structures in floorplans. To alleviate this issue, some methods \cite{min2022laser,mendez2020sedar,grader2025supercharging} utilize additional semantic information or room category annotations to assist visual FLoc. However, such semantic information requires complex manual annotation and is thus not always available. In this work, we propose an unsupervised learning technique to model scene context information, which is integrated into the visual FLoc algorithm to mitigate localization ambiguity.

\subsection{Unsupervised Visual Pre-training}

In addition to visual pre-trainings \citep{chen2020simple, du2021curious} based solely on RGB images, more and more work has proposed cross-modal pre-training methods \cite{hong2023learning,chen2024embodied,zhu2024spa} to model scene semantics and geometric priors. For example, Ego$^2$-Map \citep{hong2023learning} proposes to learn scene priors by aligning egocentric views with 2D semantic maps in a cross-modal manner. 3DLFVG \citep{zhang2024towards} achieves visual grounding by aligning the 2D geometric relations in RGB images with the spatial relations between 3D objects in the point cloud. Several other 2D-3D cross-modal methods \citep{arsomngern2023learning, chen2022self, chen2024embodied} inject 3D geometric priors into the 2D visual models by semantically or spatially aligning RGB image features with the matched point cloud sets. Although these scene priors are demonstrated to improve the accuracy of visual FLoc \cite{chen20253dp}, they fail to address the ambiguous localization at the scene level. In this paper, an unsupervised visual pre-training is proposed to model room style knowledge and provide broader scene-level contextual information for visual FLoc.

\begin{figure*}[!t]
 \centering
 \includegraphics[width=0.95\linewidth]{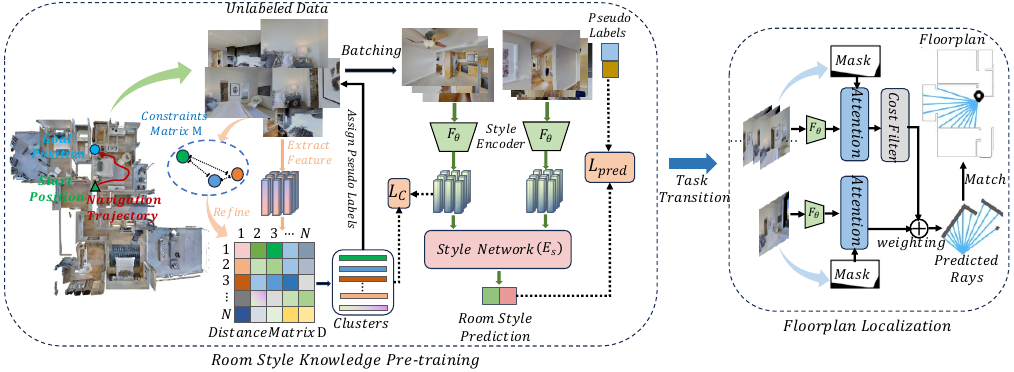}
 \caption{The left side shows the unsupervised learning technique used for room style knowledge pre-training. The unlabeled RGB images are collected automatically from indoor scene datasets based on corresponding robotic navigation episodes. We build a distance matrix $D$ based on the cosine similarity between visual features, which is further refined by a constraint matrix $M$ constructed based on the difficulty of the navigation episodes. Based on the refined distance matrix, the InfoMap clustering algorithm is used to cluster similar visual features and assign pseudo-labels to RGB images. The room discriminator is trained to predict room styles by optimizing a cluster-level contrastive loss $L_C$ and a cross-entropy loss $L_{pred}$. The well-trained room style encoder $F_\theta$ is transferred to the right side FLoc task for fine-tuning to further fit this task.}
 \label{fig2}
 \vspace{-0.3cm}
\end{figure*}

\section{Preliminaries}

\subsection{Problem Definition}

This work aims to localize RGB images to specific imaging locations in a 2D floorplan map $F$, which is represented as a matrix of dimensions $H \times W$. The floorplan is a minimalist representation of a building's layout, which retains necessary geometric occupancy information but no semantic categories. Given a time-varying image sequence $\mathcal{I}=\{\mathbf{I}_r \in \mathbb{R}^{h \times w \times 3}|r \in \{ t-k, ..., t\}\}$ containing $k+1$ RGB images, our objective is to predict the camera’s 2D location $(x, y)$ and orientation angle $\theta$ at which the image $\mathbf{I}_t$ was captured. That is, given the observation $O_{\mathcal{I}, F} = (\mathcal{I}, F)$, our goal is to infer the location parameters $S_{\mathcal{I}, F} = (x, y, \theta)$. To this end, we adopt a probabilistic framework by modeling the distribution $p(S_{\mathcal{I},F}|O_{\mathcal{I},F})$. We discretize the camera pose space as $\mathcal{S}=\{S_i\}$ and define a probabilistic map $P \in \mathbb{R}^{\hat{H} \times \hat{W} \times O}$ where each element $P(S_i)$ represents the probability $p(S_i|O_{\mathcal{I},F})$ for a candidate pose $S_i$. Here, $\hat{H}$ and $\hat{W}$ denote the number of discretized cells in the $x$ and $y$ dimensions, respectively. $O$ represents the number of orientation bins. The predicted camera pose is then given by:
\begin{flalign}
\begin{aligned}
\label{eq1}
\hat{S}_{\mathcal{I},F} = \arg\max_{S_i \in \mathcal{S}} p(S_i|O_{\mathcal{I},F})
\end{aligned}
\end{flalign}
In this work, we investigate visual FLoc in two cases: \textbf{(1)} single-frame FLoc using only the current RGB observation $\mathbf{I}_t$, i.e., $k = 0$, and \textbf{(2)} using sequential multi-frame RGB images $\mathcal{I}$ to mitigate single-frame FLoc's uncertainty and ambiguity caused by repetitive structures. 

\subsection{Background: F$^3$Loc Framework}

Our work builds upon F$^3$Loc \cite{chen2024f3loc}, a classic framework that estimates depth rays to perform visual FLoc given a single image or an image sequence. F$^3$Loc consists of a front-end observation model and a back-end Bayesian filter \citep{jonschkowski2016end}. We briefly summarize several key components of their work to provide background for our framework.

\textbf{Observation Model.} For both single-frame and multi-frame visual FLoc, F$^3$Loc uses a visual encoder $F_{\theta}$ and an attention \citep{vaswani2017attention} based network to learn the probability distribution of planar depth over a range of depth hypotheses. Specifically, given a query image or an image sequence, the observation model estimates per-column depth values that capture the distance from the camera to the nearest wall along specific angles. These values are then linearly interpolated to produce a fixed set of equiangular depth rays $\hat{r}_d \in \mathbb{R}^l$ that represent the floorplan depth, with $l$ denoting the number of predicted rays. The predicted depth rays are compared with the \textbf{G}round \textbf{T}ruth (GT) rays to calculate the likelihood scores for each grid cell and orientation, resulting in a probabilistic map $P_d \in [0,1]^{[\hat{H},\hat{W},O]}$. For each candidate location $(x,y)$ on the floorplan and each discrete orientation $\theta$, the corresponding GT rays are generated based on the floorplan's geometry.

To unify the single-frame and multi-frame settings in one framework, a selection network implemented as a multilayer perceptron is adopted to learn a weight $0 \le \omega \le 1$ from the two predictions for adaptive selection:
\begin{flalign} 
\begin{aligned}
\label{eq2}
\mathbf{P}_{fuse}=\omega Upsample(\mathbf{P}_{single}) + (1 - \omega)\mathbf{P}_{mv}.
\end{aligned}
\end{flalign}
$\mathbf{P}_{single}$ and $\mathbf{P}_{mv}$ denote the probability distributions of planar depth from a single frame and multiple frames, respectively. The upsampling operation is used to align the dimensions. The expectation of $\mathbf{P}_{fuse}$ provides the final prediction of 2D rays. $\omega$ is manually specified as 1 and 0 implying that only single-frame and multi-frame visual FLoc are used, respectively.

\textbf{Histogram Filter.} F$^3$Loc employs a histogram filter \citep{jonschkowski2016end} to keep track of the localization posterior over the entire floorplan. Such a filtering scheme is particularly effective in dealing with long-sequence FLoc.

\section{Methodology}

In this section, we first present how to pre-train a room discriminator on self-collected unlabeled room images using an unsupervised learning technique. Then, we describe how the scene layout priors summarized by the discriminator are injected into the observation model to guide visual FLoc. 

\subsection{Room Style Knowledge Pre-training}

\textbf{Data Collection.} To recognize the room type in a visual image and distinguish it from other room types, we need to train a room discriminator using images taken in different rooms. To avoid focusing on the variation of objects between rooms instead of the style of the room when summarizing the room relationships, it is also necessary to take images from different angles within the same room. By doing so, these collected images will have completely different objects while still representing the same room. In addition, images taken in different scenes should be available to ensure the universality of the room's stylistic representation. 

Considering the scarcity of indoor scene datasets annotated with room types, we propose an unsupervised learning technique to summarize scene layout priors from unlabeled scene images. We take the Gibson indoor scene dataset \cite{xia2018gibson} as an example and use the corresponding robotic navigation dataset \cite{mezghan2022memory} to collect RGB images across rooms. Specifically, for a given navigation episode $E_m$ from the training split, we first obtain the start position $p_s$ and the goal position $p_g$, as shown in Figure \ref{fig2} (left). Then, we place a robot equipped with an egocentric RGB camera at each of these two positions to capture images from different angles. Finally, we assign three category attributes to each collected image $I_i$ to record which scene it comes from ($Scene_i$), which navigation episode it comes from ($E_i$), and the difficulty of the navigation episode it belongs to ($Ed_i$). The difficulty is positively related to the length of the trajectory traversed by the navigation episode.

During the image collection process, we find that some images may contain little room style information, e.g., when the RGB camera is too close to a wall, the captured image will be completely black or white. Since using these blank images during training would provide confusing guidance, we chose to discard them. Specifically, we feed the collected images into the \textbf{S}egmentation \textbf{A}nything \textbf{M}odel (SAM) \cite{kirillov2023segment} to obtain object segmentation masks for the whole image. If the number of object masks is less than a specified threshold, the corresponding image is considered as a blank image and discarded. Overall, our data collection pipeline is automated and can provide semantic-rich RGB images for unsupervised pre-training of room style representations.

\textbf{Unsupervised Learning with Constraints.} Since the data collected from the Gibson dataset lacks room annotations, we use an unsupervised training learning technique based on a clustering algorithm to train the room discriminator, which consists of a room style encoder and a style network. We observe that the navigation episodes in the Gibson dataset can be divided into three different difficulty levels according to the trajectory length: easy (1.5-3 m), medium (3-5 m) and hard (5-10 m). Intuitively, if the start and goal positions are far apart (in a hard episode), they are likely to be in different rooms. With this assumption, we use the following rules to construct a constraint matrix $M$ of size $N \times N$ to summarize the room relationship between any two images:

(1) If $Scene_i \ne Scene_j$, then images $I_i$ and $I_j$ are not in the same room, set $M_{i,j}=-1$.

(2) If images $I_i$ and $I_j$ are taken at the same position, they are definitely in the same room, set $M_{i,j}=1$.

(3) If $E_i = E_j$ and $Ed_i = Ed_j = easy$, then images $I_i$ and $I_j$ are probably in the same room, set $M_{i,j}=0.5$.

(4) If $E_i = E_j$ and $Ed_i = Ed_j = hard$, then images $I_i$ and $I_j$ are probably in different rooms, set $M_{i,j}=-0.5$.

As shown in Figure \ref{fig2} (left), we use the above room relation constraints to achieve room style knowledge pre-training. Specifically, a ResNet50 pre-trained on ImageNet \cite{deng2009imagenet} is first used as the room style encoder to extract room images as feature vectors. We then employ the cosine similarity to measure the distance between pairs of feature vectors to construct a distance matrix $D$. The pre-built constraint matrix $M$ is used to refine the distance matrix $D$ according to the following rule:
\begin{flalign} 
\begin{aligned}
\label{eq3}
RefinedMatrix = D - \lambda M,
\end{aligned}
\end{flalign}
where $\lambda$ is a hyperparameter. Finally, based on the refined distance matrix, we use InfoMap clustering algorithm \cite{2008Maps} to cluster similar features and further assign pseudo-labels to collected RGB images. We optimize the room style encoder $F_{\theta}$ using a cluster-level contrastive loss, which is formulated as: 
\begin{flalign} 
\begin{aligned}
\label{eq4}
L_{\text{C}} = -\log \frac{\exp(F_{\theta}(I_i) \cdot \phi_+ / \tau)}{\sum_{k=1}^{K} \exp(F_{\theta}(I_i) \cdot \phi_k / \tau)},
\end{aligned}
\end{flalign}
where $K$ denotes the number of cluster representations and $\phi_k$ denotes the cluster centroid defined by the mean feature vectors of each cluster. $\phi^{+}$ denotes a cluster center which shares the same label with image $I_i$. As shown in Figure \ref{fig2} (left), images $I_i$ and $I_j$ are encoded through $F_\theta$ and then fed into a style network $E_s$ to predict whether they are taken in the same room. We use the cross-entropy loss to optimize the style network and style encoder, which is formulated as follows: 
\begin{flalign} 
\begin{aligned}
\label{eq5}
L_{\text{pred}} = -\sum_{n=1}^{N} \left[ y_i \cdot \log(E_s(F_{\theta}(I_i), F_{\theta}(I_j))) + \right. \\
\left. (1 - y_i) \cdot \log(1 - E_s(F_{\theta}(I_i), F_{\theta}(I_j))) \right],
\end{aligned}
\end{flalign}
where $y_i$ is the assigned pseudo-labels. The overall total loss can be formulated as follows:
\begin{flalign} 
\begin{aligned}
\label{eq6}
L_{loss} = L_{C} + \gamma L_{pred},
\end{aligned}
\end{flalign}
where $\gamma$ is a hyperparameter used to balance the two losses.

\subsection{Room Style Knowledge Enhanced Visual FLoc}

Through the above room style knowledge pre-training, we implicitly integrate room style knowledge into the 2D visual encoder $F_\theta$. As shown in Figure \ref{fig2} (right), the fully pre-trained room style encoder is transferred to the visual FLoc task for fine-tuning to fit the task. As described in the background section, we investigate the single-frame and multi-frame visual FLoc techniques in the F$^3$Loc framework \citep{chen2024f3loc}. F$^3$Loc localizes by finding the pose in the floorplan that has the most similar 2D rays (similar to LIDAR scans) as the prediction, as shown in Figure \ref{fig2} (right). For the training of FLoc models, we optimize an L1 loss and a cosine similarity-based shape loss:
\begin{flalign} 
\begin{aligned}
\label{eq7}
\mathcal{L}_{FLoc} = ||\mathbf{d}, \mathbf{d}^{*}||_1 + \frac{\mathbf{d}^{\top}\mathbf{d}^{*}}{max\{||\mathbf{d}||_2||\mathbf{d}^{*}||_2,\epsilon \}}.
\end{aligned}
\end{flalign}
Where $\mathbf{d}$ and $\mathbf{d}^{*}$ are predicted and GT 2D-ray depths, respectively. $\epsilon$ is a small constant to prevent division by zero.

\renewcommand\arraystretch{0.95}
\begin{table*}[!t] \small
\begin{center}
\setlength{\tabcolsep}{1.6mm}{
\begin{tabular}{l c c c c | c c c c }
\bottomrule[1.3pt]
\multirow{2}{*}{\textbf{Method \tiny{(Venue)}}} &
\multicolumn{4}{c}{\textbf{Gibson(f) R@}} & \multicolumn{4}{c}{\textbf{Gibson(g) R@}} \\
\cline{2-9}
& \textbf{0.1 m}  & \textbf{0.5 m}  & \textbf{1 m} & \textbf{1 m 30$^\circ$} & \textbf{0.1 m} & \textbf{0.5 m} & \textbf{1 m}  & \textbf{1 m 30$^\circ$} \\
\hline
PF-net\tiny{(CoRL 2018)} & 0 & 2.0 & 6.9 & 1.2 & 1.0 & 1.9 & 5.6 & 1.9 \\
MCL\tiny{(ICRA 1999)} & 1.6 & 4.9 & 12.1 & 8.2 & 2.3 & 6.2 & 9.7 & 7.3 \\
LASER\tiny{(CVPR 2022)} & 0.4 & 6.7 & 13.0 & 10.4 & 0.7 & 7.0 & 11.8 & 9.5 \\
\hline
F$^{3}$Loc$_s$\tiny{(CVPR 2024)} & 4.7 & 28.6 & 36.6 & 35.1 & 4.3 & 26.7 & 33.7 & 32.3 \\
F$^{3}$Loc$_m$\tiny{(CVPR 2024)} & 13.2 & 40.9 & 45.2 & 43.7 & 9.3 & 27.0 & 31.0 & 29.2\\ 
F$^{3}$Loc$_f$\tiny{(CVPR 2024)} & 14.3 & 42.1 & 47.4 & 45.6 & 12.2 & 39.4 & 44.5 & 43.2 \\
\hline
3DP$_s$\tiny{(MM 2025)} & 5.3 & 33.2 & 39.8 & 38.4 & 9.4 & 37.4 & 43.1 & 41.5 \\
3DP$_m$\tiny{(MM 2025)} & 15.3 & 42.5 & 47.4 & 45.9 & 11.2 & 36.3 & 41.6 & 39.8 \\
3DP$_f$\tiny{(MM 2025)} & 16.0 & 45.2 & 50.0 & 48.7 & 13.7 & 41.5 & 46.4 & 44.5 \\
\hline
Ours$_s$ & 8.3\scriptsize{(+3.0 $\uparrow$)} & 38.5\scriptsize{(+5.3 $\uparrow$)} & 45.3\scriptsize{(+5.5 $\uparrow$)} & 43.6\scriptsize{(+5.2 $\uparrow$)} & 10.7\scriptsize{(+1.3 $\uparrow$)} & 38.4\scriptsize{(+1.0 $\uparrow$)} & 44.3\scriptsize{(+1.2 $\uparrow$)} & 42.4\scriptsize{(+0.9 $\uparrow$)} \\
Ours$_m$ & 15.6\scriptsize{(+0.3 $\uparrow$)} & 44.3\scriptsize{(+1.8 $\uparrow$)} & 49.5\scriptsize{(+2.1 $\uparrow$)} & 47.7\scriptsize{(+1.8 $\uparrow$)} & 12.8\scriptsize{(+1.6 $\uparrow$)}  & 37.7\scriptsize{(+1.4 $\uparrow$)} & 44.6\scriptsize{(+3.0 $\uparrow$)} & 41.9\scriptsize{(+2.1 $\uparrow$)} \\
Ours$_f$ &  16.5\scriptsize{(+0.5 $\uparrow$)} & 47.3\scriptsize{(+2.1 $\uparrow$)} & 51.7\scriptsize{(+1.7 $\uparrow$)}  & 50.0\scriptsize{(+1.3 $\uparrow$)} &  14.3\scriptsize{(+0.6 $\uparrow$)} & 42.6\scriptsize{(+1.1 $\uparrow$)} & 48.5\scriptsize{(+2.1 $\uparrow$)} & 45.9\scriptsize{(+1.4 $\uparrow$)} \\
\bottomrule[1.3pt]
\end{tabular}}
\end{center}
\caption{Comparative studies between our single-frame ($Ours_s$), multi-frame ($Ours_m$), and adaptive ($Ours_f$) visual FLoc methods with baselines on Gibson(f) and Gibson(g) datasets.}
\label{table1}
\vspace{-0.3cm}
\end{table*}

\section{Experiments}

\subsection{Experimental Setup}

\textbf{Datasets.} We first employ a series of Gibson \citep{xia2018gibson} datasets (Gibson(g), Gibson(f), and Gibson(t)) collected by F$^{3}$Loc to fully evaluate our single-frame ($Ours_s$), multi-frame ($Ours_m$), and adaptive ($Ours_f$) visual FLoc methods. We follow the data split in F$^{3}$Loc, including 108 training scenes, 9 validation scenes, and 9 test scenes. The horizontal \textbf{F}ield \textbf{O}f \textbf{V}iew (FOV) of the images in the Gibson datasets is 108$^\circ$. The resolution of the floorplan extracted from the Gibson datasets is 0.1 m. Gibson(g) consists of general motions (including in-place steering motions) and includes 49,558 pieces of sequential views, each of which contains 4 image frames. Gibson(f) consists of only forward motions and includes 24,779 pieces of sequential views, each of which likewise contains 4 image frames. Therefore, Gibson(g) is intuitively more complex and harder than Gibson(f). Gibson(t) consists of 118 pieces of long-sequence views, each of which contains 280 $\sim$ 5152 image frames.

In addition, we use the challenging Structured3D (full) \citep{zheng2020structured3d} dataset to perform comparative studies between our single-frame FLoc method $Ours_s$ and the SoTA methods. Structured3D (full) is a photorealistic dataset containing 3296 fully furnished indoor environments with in total 78,453 perspective images. Notably, we use monocular images rather than panoramic images, and the horizontal FOV of each image is 80$^\circ$. The resolution of the floorplan extracted from the Structured3D (full) dataset is 0.02 m. For model training and evaluation, we use the official data splits.

\textbf{Baselines.} We compare our method with the following FLoc baselines, none of them using semantic labels or room category annotations. \textbf{(1) PF-net} \citep{karkus2018particle} proposes a particle filter specialized for visual FLoc. Its observation model aims to learn the similarity between an image and the corresponding map patch. \textbf{(2) MCL} \citep{dellaert1999monte} is the most popular framework for 2D localization on pure geometry maps. \textbf{(3) LASER} \citep{min2022laser} represents the floorplan as a set of points and gathers the features of the visible points of each pose in the floorplan. It actively compares the rendered pose features with the query image features for visual FLoc. \textbf{(4) F$^3$Loc} \citep{chen2024f3loc} is a classic visual FLoc method that proposes a probabilistic model consisting of a ray-based observation module and a histogram filtering module. F$^3$Loc includes three variants: single-frame (F$^3$Loc$_s$), multi-frame (F$^3$Loc$_m$), and adaptive (F$^3$Loc$_f$) visual FLoc methods. \textbf{(5) 3DP} \citep{chen20253dp} is one of the SoTA visual FLoc methods. 3DP injects 3D geometric priors into visual FLoc which significantly improve the single-frame and multi-frame FLoc accuracy without the need of any semantic labels. 3DP also includes three variants: single-frame (3DP$_s$), multi-frame (3DP$_m$), and adaptive (3DP$_f$) visual FLoc methods. 


\textbf{Metrics.} Following existing work \cite{chen2024f3loc, chen20253dp}, we report recall metrics computed at localization accuracies of 0.1 m, 0.5 m, and 1 m. We also report recall for predictions with an orientation error bounded to less than 30$^\circ$ (with a localization accuracy of 1 m). Recall is calculated as the percentage of predictions that fall within these thresholds. For comparisons on the Gibson(t) dataset, the \textbf{R}oot-\textbf{M}ean-\textbf{S}quare \textbf{E}rror (RMSE) is also employed to measure the accuracy of sequential trajectory tracking when localization is successful (RMSE(S)) and in all cases (RMSE(A)).

\begin{table}[!t] \small
\begin{center}
\setlength{\tabcolsep}{0.4mm}{
\begin{tabular}{l | c c c c }
\bottomrule[1.3pt]
\multirow{2}{*}{\textbf{Method \tiny{(Venue)}}} &
\multicolumn{4}{c}{\textbf{Gibson(t)}} \\
\cline{2-5}
& \textbf{R@0.2 m} & \textbf{R@1 m} & \textbf{RMSE(S)} & \textbf{RMSE(A)} \\
\hline
LASER\tiny{(CVPR 2022)} & - & 59.5 & 0.39 & 1.96 \\
F$^{3}$Loc$_s$\tiny{(CVPR 2024)} & 35.1 & 89.2 & 0.18 & 0.88 \\
3DP$_s$\tiny{(MM 2025)} & 54.1 & 89.2 & 0.16 & 0.75 \\
\hline
Ours$_s$ & 67.6\scriptsize{(+13.5 $\uparrow$)} & 94.6\scriptsize{(+5.4 $\uparrow$)} & 0.13\scriptsize{(-0.03 $\downarrow$)} & 0.51\scriptsize{(-0.24 $\downarrow$)} \\
\bottomrule[1.3pt]
\end{tabular}}
\end{center}
\caption{Comparative studies of long-sequence visual Floc methods on the Gibson(t) dataset.}
\label{table2}
\vspace{-0.3cm}
\end{table}


\textbf{Implementation Details.} For the room style knowledge pre-training, we only collect RGB images from the training split of the Gibson dataset \cite{xia2018gibson} to prevent data leakage. The height and radius of the robot used to collect data are 1.5 m and 0.1 m, respectively. The robot has a single RGB sensor with a 90$^\circ$ FOV. We train the style encoder and style network for 20 epochs using the Adam optimizer \cite{kingma2014adam} with a weight decay of 5e-4 and a batch size of 64. We set the refinement hyperparameter $\lambda$ as a learnable parameter. The balance parameter $\gamma$ in Equation (6) is set to 1.0. 

For the fine-tuning of visual FLoc, we use an Adam optimizer with a learning rate of 0.001 for all training. For the Structured3D (full) dataset, the single-frame FLoc model $Ours_s$ is trained for 100 epochs. For the Gibson(f) and Gibson(g) datasets, $Ours_s$, $Ours_m$, and $Ours_f$ are trained on the entire training split for 100, 100, and 20 epochs, respectively. During the training of $Ours_f$, the parameters of single-frame and multi-frame FLoc methods are frozen. Only the selection network is trained by adopting the training paradigm in \cite{chen20253dp} to prevent biased learning. All model training is performed on 4 NVIDIA 3090 GPUs. Follow the settings in \cite{chen2024f3loc}, Our single-frame and multi-frame FLoc methods match the predicted 40 and 160 rays to the floorplans for localization, respectively. The cost filter in Figure \ref{fig2} (right) is implemented as a UNet-style \citep{ronneberger2015u} network that converts multi-channel image features into single-channel features for ray prediction. The selection network is implemented as 3 stacked linear layers with BatchNorm \citep{ioffe2015batch} and ReLU activation.


\begin{table}[!t] \small
\begin{center}
\setlength{\tabcolsep}{0.8mm}{
\begin{tabular}{l | c c c c }
\bottomrule[1.3pt]
\multirow{2}{*}{\textbf{Method \tiny{(Venue)}}} &
\multicolumn{4}{c}{\textbf{Structured3D (full) R@}} \\
\cline{2-5}
& \textbf{0.1 m}  & \textbf{0.5 m}  & \textbf{1 m}  & \textbf{1 m 30$^\circ$} \\
\hline
PF-net\tiny{(CoRL 2018)} & 0.2 & 1.3 & 3.2 & 0.9  \\
MCL\tiny{(ICRA 1999)} & 1.3 & 5.2 & 7.8 & 6.4 \\
LASER\tiny{(CVPR 2022)} & 0.7 & 6.4 & 10.4 & 8.7 \\
F$^{3}$Loc$_s$\tiny{(CVPR 2024)} & 1.5 & 14.6 & 22.4 & 21.3 \\
3DP$_s$\tiny{(MM 2025)} & 5.6 & 27.4 & 55.5 & 24.0 \\
\hline
Ours$_s$ & 6.4\scriptsize{(+0.8 $\uparrow$)} & 28.6\scriptsize{(+1.2 $\uparrow$)} & 56.9\scriptsize{(+1.4 $\uparrow$)} & 25.2\scriptsize{(+1.2 $\uparrow$)} \\
Oracle & 61.0 & 93.8 & 94.9 & 94.6 \\
\bottomrule[1.3pt]
\end{tabular}}
\end{center}
\caption{Comparative studies of single-frame visual FLoc methods on the Structured3D (full) dataset.}
\label{table3}
\vspace{-0.3cm}
\end{table}

\subsection{Comparative studies with SoTA methods}


We first compare our method with the baseline and SoTA methods on the Gibson (f) and Gibson (g) datasets, the results are shown in Table \ref{table1}. Existing FLoc methods are severely affected by repetitive structures in the floorplans, which can easily lead to ambiguous or even incorrect localization. Thanks to room style knowledge, our method can effectively mitigate localization uncertainty. Quantitatively, our single-frame FLoc method $Ours_s$ improves the four metrics by 3.0\%, 5.3\%, 5.5\%, and 5.2\% relative to 3DP$_s$ on Gibson(f), respectively. Even though the Gibson (g) dataset is more challenging, $Ours_s$ can still improve the FLoc accuracy. Notably, $Ours_s$ achieves comparable FLoc performance on both datasets, reflecting that our method can deal with drastic visual changes caused by in-place turning. Since our method can identify and distinguish room types, the performance improvement of visual FLoc extends to multi-frame and adaptive FLoc methods. As shown in Table \ref{table1}, $Ours_m$ and $Ours_f$ significantly improve FLoc accuracy compared to 3DP. In addition, the performance gains of $Ours_m$ on the more challenging Gibson(g) are greater than those on Gibson(f). Quantitatively, our multi-frame FLoc method $Ours_m$ improves the four metrics by 1.6\%, 1.4\%, 3.0\%, and 2.1\% relative to 3DP$_m$ on Gibson(g), respectively. As expected, the adaptive method $Ours_f$ achieves the best performance on both datasets by integrating the strengths of single-frame and multi-frame FLoc methods.

\begin{figure}[!t]
 \centering
 \includegraphics[width=1.0\linewidth]{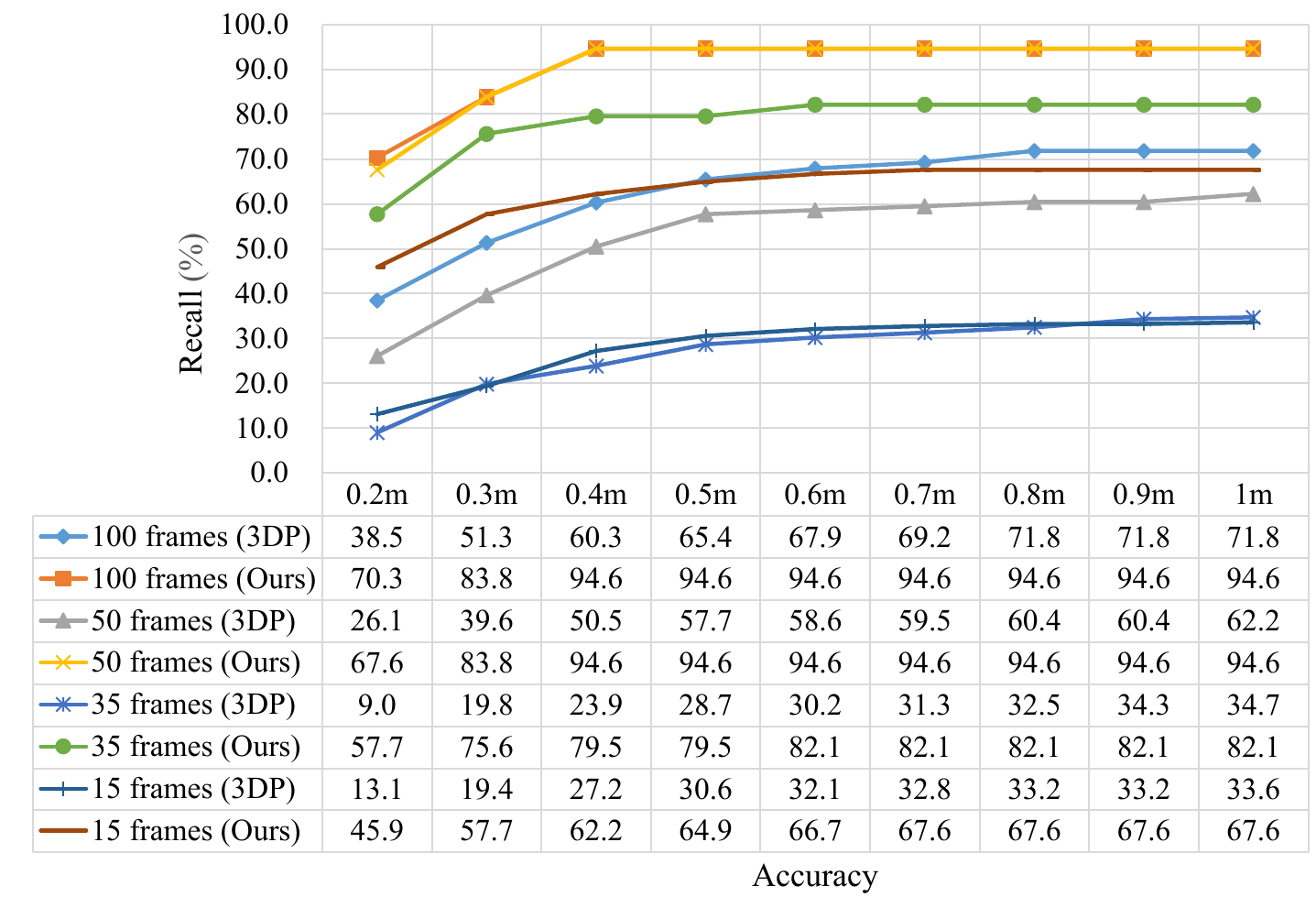}
 \caption{Comparison of the localization performance using different number of historical frames. The more frames are used within the filter, the higher the localization success.}
 \label{fig3}
 \vspace{-0.3cm}
\end{figure}

We employ well-trained visual FLoc models to solve the long-sequence trajectory tracking problem on the Gibson(t) dataset, the results are shown in Table \ref{table2}. Technically, we combine the histogram filter proposed by F$^3$Loc with our room style Knowledge enhanced observation model. Notably, our single-frame FLoc method improves the recall metric by 13.5\% compared to 3DP$_s$ at a localization accuracy of 0.2 m. In addition, the reduction in the RMSE metrics reflects the robustness of our method in sequential trajectory tracking. In addition, we conduct comparative studies between $Ours_s$ and baselines on the Structured3D (full) dataset, the results are shown in Table \ref{table3}. Although Structured3D (full) dataset is challenging due to its varied scenes (across 3296 scenes), our method can still improve FLoc performance at various accuracy levels. Such performance gains imply that identifying room styles helps to clearly locate visual observations in floorplans. In addition, we report the FLoc performance using GT depth information (Oracle), reflecting the potential of schemes that match depth-based geometric cues. Figure \ref{fig3} illustrates the changes in recall versus accuracy for long-sequence trajectory tracking using different numbers of historical frames. Our method outperforms 3DP in all localization accuracies.

\begin{table}[t] \small
\begin{center}
\setlength{\tabcolsep}{1.9mm}{
\begin{tabular}{c c | c c c c}
\bottomrule[1.3pt]
\multicolumn{2}{c}{\textbf{Ablations}}&
\multicolumn{4}{c}{\textbf{Gibson(g) R@}}\\
\hline
Data Clean & Dist Refine & \textbf{0.1 m}  & \textbf{0.5 m}  & \textbf{1 m}  & \textbf{1 m 30$^\circ$} \\
\hline
\checkmark &  & 9.4 & 34.7 & 42.1 & 39.5 \\
& \checkmark & 9.8 & 36.8 & 43.0 & 41.2 \\
\checkmark & \checkmark & 10.7 & 38.4 & 44.3 & 42.4 \\
\bottomrule[1.3pt]
\end{tabular}}
\end{center}
\caption{Ablation studies on data cleaning and distance refinement.}
\label{table5}
\vspace{-0.3cm}
\end{table}

\begin{table}[!t] \small
\begin{center}
\setlength{\tabcolsep}{2.4mm}{
\begin{tabular}{l | c c c c }
\bottomrule[1.3pt]
\multirow{2}{*}{\textbf{Method \tiny{(Venue)}}} &
\multicolumn{4}{c}{\textbf{Gibson(g) R@}} \\
\cline{2-5}
& \textbf{0.1 m}  & \textbf{0.5 m}  & \textbf{1 m}  & \textbf{1 m 30$^\circ$}\\
\hline
SimCLR\tiny{(ICML 2020)} & 4.7 & 28.2 & 35.3 & 34.6 \\
CRL\tiny{(ICCV 2021)} & 5.0 & 29.7 & 37.2 & 35.8 \\
Ego$^2$-MAP\tiny{(ICCV 2023)} & 5.7 & 30.6 & 36.9 & 35.2 \\
ECL\tiny{(MM 2024)} & 7.1 & 34.8 & 40.5 & 38.7 \\
SPA\tiny{(ICLR 2025)} & 8.3 & 35.7 & 41.4 & 39.5 \\
3DP\tiny{(MM 2025)} & 9.4 & 37.4 & 43.1 & 41.5 \\
\hline
Ours & 10.7 & 38.4 & 44.3 & 42.4 \\
\bottomrule[1.3pt]
\end{tabular}}
\end{center}
\caption{Comparative studies of enhancing visual Floc by using different unsupervised visual pre-trainings.}
\label{table6}
\vspace{-0.3cm}
\end{table}

\subsection{Ablation Study}

The ablation of room style knowledge pre-training is equivalent to rolling back our method back to F$^3$Loc. The significant performance gains of our method relative to F$^3$Loc highlight the effectiveness of integrating room style knowledge into visual FLoc algorithms. In addition, we perform ablation studies on data cleaning during data collection and distance refinement during pre-training, as shown in Table \ref{table5}. Results on Gibson (g) indicate that both components improve FLoc performance at various localization accuracies.

As shown in Table \ref{table6}, our room style knowledge pre-training is compared with existing unsupervised visual pre-trainings to demonstrate the superiority of our method. SimCLR and CRL are contrastive pre-trainings based on pure RGB images. Ego$^2$-MAP, ECL, SPA, and 3DP are all cross-modal visual pre-trainings that use 2D/3D scene priors to enhance visual features. Technically, these comparisons are achieved by integrating their pre-trained visual encoders into the visual FLoc algorithm. The results in Table \ref{table6} emphasize the advantages of our scene context modeling compared to other visual pre-training methods.

\begin{figure}[!t]
 \centering
 \includegraphics[width=1.0\linewidth]{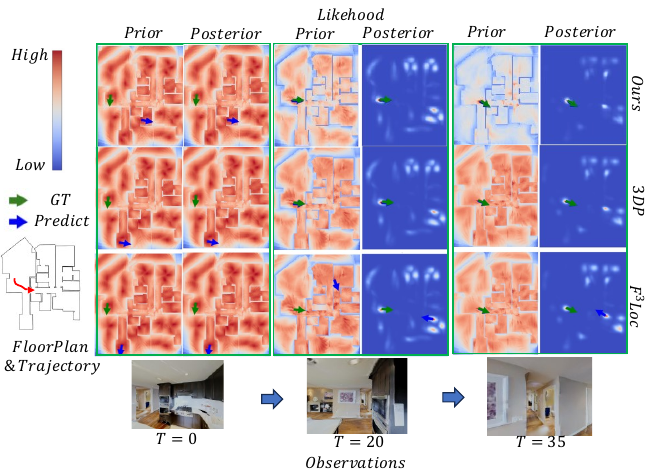}
 \caption{Qualitative comparative studies with F$^3$Loc and 3DP in a scene with complex furniture.}
 \label{fig5}
 \vspace{-0.3cm}
\end{figure}

\subsection{Qualitative Evaluation}

As shown in Figure \ref{fig5}, our method is compared qualitatively with F$^3$Loc and 3DP in a scene with complex furniture. Compared with F$^3$Loc, our single-frame localization is highly consistent with GT in most cases, as shown in the prior probabilistic maps. As shown in the posterior probabilistic maps, our long-sequence trajectory tracking quickly syncs the predicted pose to the same modality with GT, whereas F$^3$Loc shows long-term localization errors. Benefiting from our room style knowledge pre-training, the prior probability maps at $T=20$ and $T=35$ reflect that our single-frame FLoc method is more deterministic and can effectively alleviate localization ambiguity.

\section{Conclusion and Future Work}

In this paper, we propose using broader scene context information to mitigate ambiguous visual FLoc, rather than relying solely on 2D/3D geometric cues. In particular, we propose an unsupervised training technique based on a clustering algorithm to train a room discriminator. The room discriminator can identify and distinguish various room types, whose room style knowledge is injected into the visual FLoc algorithm to guide definite FLoc. Comparative and ablation studies on two standard visual FLoc benchmarks demonstrate the superiority of our method. Notably, this work emphasizes fully utilizing the room style knowledge for visual FLoc, which does not mean that geometric cues are not important. In future work, we will explore the advantages of integrating 3D geometric priors and scene context information by proposing a unified visual FLoc framework.

\section{Acknowledges}
This work was supported in part by the National Natural Science Foundation of China under 62272489, 62332020, and 62350004, in part by the Natural Resources Science and Technology Plan Project of Hunan Province under 2021-17, and in part by the Open Competition Project of Xiangjiang Laboratory under 23XJ01011. This work was carried out in part using computing resources at the High-Performance Computing Center of Central South University.

\bibliography{aaai2026}

\end{document}